\documentclass[10pt,twocolumn,letterpaper]{article}

\usepackage{cvpr}              

\usepackage{algorithm}
\usepackage{algorithmic}
\usepackage{graphicx}
\usepackage{amsmath}
\usepackage{amssymb}
\usepackage{booktabs}
\usepackage{amsfonts}

\usepackage{multirow}
\usepackage{array}
\usepackage{color}
\usepackage{subcaption}
\usepackage{colortbl}
\usepackage{arydshln}
\graphicspath{{images/}}
\usepackage{bbding}
\usepackage{makecell}
\usepackage{hhline}
\usepackage{bm}

\usepackage[pagebackref,breaklinks,colorlinks]{hyperref}

\usepackage[capitalize]{cleveref}
\crefname{section}{Sec.}{Secs.}
\Crefname{section}{Section}{Sections}
\Crefname{table}{Table}{Tables}
\crefname{table}{Tab.}{Tabs.}

\def\cvprPaperID{*****} 
\def\confName{CVPR}
\def\confYear{2022}

\begin{document}

\title{1st Place Solution to ECCV 2022 Challenge on Out of Vocabulary Scene Text Understanding: End-to-End Recognition of Out of Vocabulary Words}

\author{Zhangzi Zhu, Chuhui Xue, Yu Hao, Wenqing Zhang, Song Bai\\
ByteDance Inc.
}
\maketitle

\begin{abstract}
Scene text recognition has attracted increasing interest in recent years due to its wide range of applications in multilingual translation, autonomous driving, etc. In this report, we describe our solution to the Out of Vocabulary Scene Text Understanding (OOV-ST) Challenge, which aims to extract out-of-vocabulary (OOV) words from natural scene images. Our oCLIP-based model achieves 28.59\% in h-mean which ranks 1st in end-to-end OOV word recognition track of OOV Challenge in ECCV2022 TiE Workshop. 
\end{abstract}

\section{Introduction}
Recently, many scene text recognition techniques have been proposed which learns the language knowledge for better recognition the in-vocabulary (IV) words (\textit{i.e.}, the words have been appeared in the training set). However, the language information of out-of-vocabulary (OOV) words are usually difficult to learn if they have never been seen during training, which makes model difficult to recognize the OOV words accurately.

ECCV 2022 Challenge on Out of Vocabulary Scene Text Understanding \footnote{\scriptsize\url{https://rrc.cvc.uab.es/?ch=19}}, held together with ECCV 2022 workshop on Text in Everything (TiE) \footnote{\scriptsize\url{https://sites.google.com/view/tie-eccv2022/challenge}},  aims to evaluate the model performances on recognizing OOV words. In this challenge, the training, validation and test sets are composed of several commonly used datasets, including ICDAR13 \cite{karatzas2013icdar}, ICDAR15 \cite{karatzas2015icdar}, MLT19 \cite{nayef2019icdar2019}, COCO-Text \cite{veit2016coco}, TextOCR \cite{singh2021textocr}, HierText \cite{long2022towards}, and OpenImagesText \cite{krylov2021open}. Two evaluation metrics are provided that focuses on: (1) OOV words only which aims to evaluate the model performances on recognizing the OOV words; and (2) both IV and OOV words by averaging the IV and OOV scores which aims to consider both IV and OOV words in evaluation.

In this report, we present our solution to the end-to-end OOV word recognition task. We first pre-train different commonly-used network backbones by using oCLIP \cite{xue2022language}. We then fine-tune PAN \cite{wang2019efficient}, Mask TextSpotter-v3 (MTS-v3) \cite{liao2020mask} and TESTR \cite{Zhang_2022_CVPR} on the composed datasets for word detection. Finally, we recognize the detected words by using a SCATTER-based \cite{litman2020scatter} recognizer. Our method achieves 28.59\% h-mean, which ranks 1st on end-to-end recognition.

\section{Methods and Experimental Results}

\subsection{Text Detection}
In our solution, we first pre-train different backbones including VAN-large \cite{guo2022visual} and Deformable ResNet-101 \cite{dai2017deformable} by using oCLIP \cite{xue2022language}. Next, we fine-tune PAN \cite{wang2019efficient}, MTS-v3 \cite{liao2020mask} and TESTR \cite{Zhang_2022_CVPR} on the composed datasets by using the pre-trained models. Finally, we combine the detection results from different models together and recognize the words following \cite{zhu2022oovrec}. All models are evaluated on the validation set of the composed dataset.

\subsubsection{Model Pre-train}
We first pre-train VAN-large \cite{guo2022visual} and Deformable ResNet-101 \cite{dai2017deformable} by using oCLIP \cite{xue2022language} on SynthText \cite{Gupta16} dataset as well as the provided composed dataset. Next, we fine-tune PAN \cite{wang2019efficient} (with VAN-large as backbone), MTS-v3 \cite{liao2020mask} (with Deformable ResNet-101 as backbone), and TESTR \cite{Zhang_2022_CVPR} (with Deformable ResNet-101 as backbone) on the composed dataset by using the pre-trained backbone weights. By pre-training using oCLIP, the performances of different models have been improved by 1\% to 3\% in Fscore as shown in Table \ref{tab:oclip}. 

\begin{table*}[!h]
\centering
\begin{tabular}{lcclccl}
\toprule
\multirow{2}{*}{\textbf{Method}} & \multicolumn{3}{c}{\textbf{OOV}} & \multicolumn{3}{c}{\textbf{All}} \\
\cmidrule(lr){2-4}  \cmidrule(l){5-7}
& \textbf{Precision} & \textbf{Recall} & \textbf{Fscore} & \textbf{Precision} & \textbf{Recall} & \textbf{Fscore} \\
\midrule
PAN                     & 65.36  & 68.71  & 67.00 & 83.36  & 56.18  & 67.21 \\
PAN+oCLIP               & 64.03  & 73.11  & \textbf{68.27 (+1.27)} & 83.37  & 61.64  & \textbf{70.88 (+3.67)} \\
MTS-v3                     & 77.13  & 48.31  & 59.41 & 87.61  & 42.16  & 56.93 \\
MTS-v3+oCLIP               & 77.55  & 48.83  & \textbf{59.93 (+0.52)} & 87.73  & 43.09  & \textbf{57.87 (+0.94)} \\
TESTR                   & 69.55  & 55.12  & 61.50 & 84.75  & 52.34  & 64.71 \\
TESTR+oCLIP             & 71.47  & 56.22  & \textbf{62.93 (+1.43)} & 85.93  & 55.83  & \textbf{65.73 (+1.02)} \\
\bottomrule
\end{tabular}
\caption{Text detection results by adopting oCLIP\cite{xue2022language} for backbone pre-training.}
\label{tab:oclip}
\end{table*}

\subsubsection{Model Ensemble}
Next, we collect the detection results from different models with different scales of images (\textit{i.e.}, 512, 960, 1280, 1600) as input which are hence combined together. We further apply soft-nms \cite{bodla2017soft} on the combined results and filter the detected boxes by a threshold of 0.92. Table \ref{tab:ensemble} shows the model ensemble results.

\begin{table}[!t]
\centering
\begin{tabular}{cccc}
\toprule
\multirow{2}{*}{\textbf{Method}} & \multicolumn{3}{c}{\textbf{OOV}}                       \\  
 \cmidrule(l){2-4}                                & \textbf{Precision} & \textbf{Recall} & \textbf{Fscore} \\
\midrule
PAN                              & 64.03  & 73.11  & 68.27                 \\
MTS-v3                           & 77.55  & 48.83  & 59.93                 \\
TESTR                            & 71.47  & 56.22  & 62.93                 \\
\midrule
Ensemble                     & 69.85              & 76.20           & 72.89           \\
\bottomrule
\end{tabular}
\caption{Text detection results by model ensemble.}
\label{tab:ensemble}
\end{table}

\begin{table}[!h]
\centering
\begin{tabular}{cccc}
\toprule
\multirow{2}{*}{\textbf{Set}} & \multicolumn{3}{c}{\textbf{OOV}}                       \\
\cmidrule(l){2-4}                                 & \textbf{Precision} & \textbf{Recall} & \textbf{Fscore} \\
\midrule
Validation         & 41.08              & 41.73           & 41.40          \\          
Test                      & 20.28              & 48.42           & 28.59           \\

\bottomrule
\end{tabular}
\caption{End-to-end recognition results on validation and test set, respectively.}
\label{tab:e2e}
\end{table}

\subsection{End-to-End Word Recognition}
We pass the detected texts to our recognition model \cite{zhu2022oovrec} and filter out the words that are recognized to be `ignore' texts to obtain the text recognition results. Table \ref{tab:e2e} shows the end-to-end recognition results from our models. 

\section{Conclusion}
This report presents our solutions to the end-to-end OOV word recognition task of ECCV 2022 Challenge on OOV-ST. We adopt oCLIP for model pre-train and model ensemble for better detection of texts in scenes. The presented solution ranks first in the end-to-end recognition of out of vocabulary words in the ECCV 2022 Challenges on OOV-ST.

{\small
\bibliographystyle{ieee_fullname}
\bibliography{egbib}
}

\end{document}